\documentclass[11pt,letterpaper]{article}
\usepackage{cogsys}
\usepackage[T1]{fontenc}
\usepackage{times}
\usepackage[pdftex]{graphicx} 
\usepackage{booktabs}
\usepackage{makecell}
\usepackage{amsmath}
\usepackage{algorithm}
\usepackage{algpseudocode}

\usepackage{natbib}
\newcounter{rpostulate}
\newcounter{opostulate}
\newcounter{ppostulate}
\newcounter{lpostulate}

\newenvironment{rpost}{%
  \stepcounter{rpostulate}%
  \par\noindent
  \hangindent=21pt\hangafter=1
  \textbf{R\therpostulate.}~\ignorespaces
  \setlength{\parindent}{0pt}
  \setlength{\parskip}{4pt}%
}{%
  \par\smallskip
}

\newenvironment{opost}{%
  \stepcounter{opostulate}%
  \par\noindent
  \hangindent=21pt\hangafter=1
  \textbf{O\theopostulate.}~\ignorespaces
  \setlength{\parindent}{0pt}
  \setlength{\parskip}{4pt}%
}{%
  \par\smallskip
}

\newenvironment{ppost}{%
  \stepcounter{ppostulate}%
  \par\noindent
  \hangindent=21pt\hangafter=1%
  \textbf{P\theppostulate.}~\ignorespaces
  \setlength{\parindent}{0pt}%
  \setlength{\parskip}{4pt}%
}{%
  \par\smallskip
}

\newenvironment{lpost}{%
  \stepcounter{lpostulate}%
  \par\noindent
  \hangindent=21pt\hangafter=1%
  \textbf{L\thelpostulate.}~\ignorespaces
  \setlength{\parindent}{0pt}%
  \setlength{\parskip}{4pt}%
}{%
  \par\smallskip
}

\def\trellis/{{\sc Trellis}}

\def\texitem#1{\par\noindent\hangindent 20pt
               \hbox to 20pt {\hss #1 ~}\ignorespaces}
\def\itemitem#1{\par\noindent\hangindent 40pt
                \hbox to 40pt {\hss #1 ~}\ignorespaces}
\def\bull{\texitem{$\bullet$}}

\cogsysheading{X}{2026}{1--20}{8/2026}{10/2026}

\ShortHeadings{A Unified Account of Concepts and Chunks}
              {K.\ Singaravadivelan and P.\ Langley}

\begin{document}

\title{A Unified Account of Concepts and Chunks}

\author{Karthik Singaravadivelan}{ksingara3@gatech.edu}
\author{Pat Langley}{patrick.w.langley@gmail.com}
\address{Georgia Institute of Technology, Atlanta, GA 30332 USA \\
         Institute for the Study of Learning and Expertise,
         Palo Alto, CA 94306 USA}

\begin{abstract}
Cognitive psychology has studied how people encode, use, and learn
concepts that describe categories, and how they represent, recognize,
and acquire chunks for familiar patterns of elements. The literatures
on these two topics are nearly disjoint, which poses a challenge for
unified theories of cognition. In this paper, we review Cobweb, a
computational account of categorization and concept formation, then 
propose an extended theory that incorporates chunks and their
acquisition. The theory makes no commitments about
modality, applying to any experience that decomposes into elements
and relations among them. We also present \trellis/, an
implementation of this theory, and illustrate its application to
learning context-free grammars, which we adopt as a testbed because
they involve both concept-like and chunk-like elements. In addition,
we report experimental results on learning for three synthetic grammars 
that demonstrate the system's ability to represent syntactic knowledge, 
use it to parse and generate sentences, and learn compositional 
structures from sample parses. We conclude by discussing related 
work on concepts and chunks, along with directions for future 
research on the problem. 
\end{abstract} 

\section{Introduction}

A primary aim of science is to develop general theories that explain
as broad a range of phenomena as possible. Within cognitive systems,
such theories involve postulates about the mental structures stored in
memory, the performance processes that operate over them, and the
learning mechanisms that acquire them.
Different paradigms take differing positions on these issues, but all
share a concern with generality, seeking broad coverage from a small 
set of assumptions.

In this paper, we propose a unified computational account of two
facets of cognition that researchers have long treated as distinct and
disconnected. One of these revolves around {\it concepts\/}, which
\citet{langley-concepts-chunks} defines as:

\vskip 0.05in

\bull
  {\it A concept is a cognitive structure that denotes a set of                 
       entities or situations and that characterizes them in terms
       of their regularities\/}.

\vskip 0.05in
\noindent
The representation, use, and learning of concepts has been a major
topic in cognitive psychology \citep{rips-smith-medin} and artificial
intelligence \citep{langley-machine-learning} since the 1960s. A second
area, less popular but one that still receives substantial attention,
concerns {\it chunks}, which Langley defines as:

\vskip 0.05in

\bull
  {\it A chunk is a cognitive structure that denotes a collection of 
       entities, possibly typed, that form a specified configuration 
       or relational pattern\/}.

\vskip 0.05in
\noindent
Treatments of chunk encoding, recognition, and acquisition have mainly
appeared in the psychological literature
\citep{miller-magical-seven,fountain-doyle-chunking}, but there has
been some AI work \citep{campbell-berliner-chunks}. Topics like concept generalization and similarity have been recurring themes in work on 
concepts, whereas research on chunks has emphasized the composition 
of higher-level structures from their constituents.

%
%
%

Concepts and chunks have much in common, but work on the former has
emphasized generalization, whereas studies of the latter have focused
on composition. Both capabilities are necessary for robust cognition, 
but neither alone is sufficient, which suggests that most existing 
theories are incomplete. A system that supports generalization 
without composition is limited to encoding knowledge at a single level, 
whereas one that acquires configurations without generalizing them 
cannot use them in novel situations. Both capacities are needed.  
Composable chunks supply parts and relations among them, whereas 
generalized concepts extend their coverage. 

In the remaining pages, we present a unified theory
that addresses both aspects of cognition. Our approach builds on Cobweb
\citep{fisher-cobweb}, a computational account of concept representation,
organization, retrieval, and formation. We start by reviewing Cobweb's
main assumptions, which our framework retains, and then introduce
postulates that support chunk-related phenomena. After this, we
describe \trellis/, an implementation of the theory, and report
experimental results on grammar induction. We focus on this task  
because syntax acquisition involves both generalization and composition, 
but the theory should apply equally well to vision, problem solving, 
and other cognitive functions. In closing, we examine related ideas
and note avenues for future research.




\section{The Cobweb Framework}

\citeauthor{fisher-cobweb}'s (\citeyear{fisher-cobweb}) Cobweb was an innovative model of categorization and
concept formation that combined ideas from decision-tree induction,
Bayesian classifiers, and nearest-neighbor methods. The system
received considerable attention within the maturing machine learning
community and made contact with important psychological findings
like basic-level and typicality effects. Excitement about the approach
led to multiple extensions, many of them documented in Fisher, Pazzani, 
and Langley (\citeyear{fisher-concept-formation-vol}). In this 
section, we review the framework's main theoretical tenets, which 
it shares with its descendants, including the new system we report 
later. 

\subsection{Representation in Cobweb}

The Cobweb framework's most basic postulates concern how conceptual
knowledge is represented in long-term memory, as later assumptions
build directly upon them. These include statements that:

\vskip 0.05in

\begin{rpost}
Long-term memory contains two types of structures: instances and
concepts.
\end{rpost}

\begin{rpost}
An instance is a set of attribute-value pairs that describe a single
observation.
\end{rpost}

\begin{rpost}
A concept specifies, for each attribute, a probability distribution
over that attribute's values.
\end{rpost}

\vskip 0.02in
\noindent
That is, we can view each concept as a probabilistic generalization
for a set of instances in which each attribute's distribution is
independent given the concept. Unlike many other theories of
categorization, Cobweb retains both types of structures in memory
and we can view each instance as a very specific concept with a
heavily skewed distribution.




\subsection{Organization in Cobweb}

Of course, the contents of long-term memory must be organized in
some fashion, so the Cobweb framework also takes strong positions
about their arrangement. These include postulates that:


\vskip 0.05in

\begin{opost}
Concepts reside in a taxonomic hierarchy with a single root and
instances as terminal nodes.
\end{opost}

\begin{opost}
Each nonterminal node is a probabilistic summary of the instances
below it in the taxonomy.
\end{opost}

\begin{opost}
Each nonterminal node has two or more children that partition the
instances it summarizes.
\end{opost}

\vskip 0.02in
\noindent
This organization resembles that produced by hierarchical clustering 
systems, except that each nonterminal node (cluster) in the taxonomy 
has a concept description rather than a set of instances, which are 
stored at terminal nodes.

The taxonomy is also similar to a decision tree, except that
it stores `tests' on nodes rather than links. Also, the tests
are multivariate in that each child describes distributions
over a set of attributes rather than just one. Like a decision
tree, a Cobweb hierarchy partitions instances into mutually
exclusive sets, but each node is a probabilistic generalization
of instances rather than a logical characterization. We can
also view paths through the taxonomy as {\it indexing\/} nodes
in long-term memory, which has implications for performance.




\subsection{Performance in Cobweb}

Claims about representation and organization are not enough for a
complete theory. Cobweb also posits performance mechanisms that
use its concepts, and their arrangement in long-term memory, to
produce behavior. We can summarize the framework's assumptions on
this front as:

\vskip 0.05in

\begin{ppost}
There are two linked performance mechanisms: {\it classification\/}
and {\it prediction\/}.
\end{ppost}

\vskip 0.02in
\noindent
The first assigns a case to a category, whereas the second infers
any values for missing attributes. This is similar to the distinction
in the psychological literature between recognition and recall.

\vskip 0.05in

\begin{ppost}
Classification involves sorting an instance downward through the taxonomy.
\end{ppost}

\vskip 0.02in
\noindent
Thus, classification is much like that in decision trees \citep{quinlan-id3},
except that Cobweb selects the best child at each level. The criterion
it uses is {\it category utility} \citep{fisher-cobweb}, a probabilistic
measure that favors children with higher intra-category similarity and
lower between-category similarity. The process steps greedily down the
tree, selecting the best option at each level.\footnote{Some variants,
like Cobweb/4V \citep{cobweb4v-barari}, replace this greedy descent with 
best-first search, but we will adopt the more basic scheme here, as in 
the original version of Cobweb.} 

\vskip 0.05in

\begin{ppost}
Sorting continues until reaching a terminal node or no child is better
than the current node.
\end{ppost}

\vskip 0.02in
\noindent
This is another important difference from decision trees, which always
sort cases to terminal nodes. In contrast, Cobweb halts at a nonterminal
node when none of its children have a better score.

\vskip 0.05in

\begin{ppost}
At the halting node, prediction imputes the most probable values for
unknown attributes.
\end{ppost}

\vskip 0.02in
\noindent
Again, this scheme resembles that in decision trees, but rather than
predicting a class label, Cobweb infers values for all missing
attributes, an ability that \citet{fisher-cobweb} called {\it 
flexible prediction\/}.

\subsection{Learning in Cobweb}

Finally, the theory makes statements about how conceptual knowledge
is acquired through learning. These additional postulates include:


\vskip 0.05in

\begin{lpost}
Learning is an incremental, online process that is fully interleaved
with performance.
\end{lpost}

\begin{lpost}
Learning is unsupervised in that instances do not come with class labels.
\end{lpost}

\vskip 0.02in
\noindent
That is, sorting each new instance changes both the content of
existing concepts and their organization in long-term memory. 
Moreover, the learner must invent its own categories rather 
than obtain them from an outside source. Training cases can 
include a `class' attribute, but this feature is treated no 
differently from others. 

Furthermore, the Cobweb theory posits two varieties of incremental
learning, one statistical in character and the other structural in
nature:

\vskip 0.05in

\begin{lpost}
Sorting an instance through a node in the taxonomy updates its
associated probabilities.
\end{lpost}

\vskip 0.02in
\noindent
This mechanism simply alter counts that reflect the prior probability
of a concept given its parent and the probability of each attribute
value given that concept, much as in naive Bayesian classifiers.
These changes are determined entirely by the instance's content
and the path it traverses.

\vskip 0.05in

\begin{lpost}
At each node, sorting considers four structural options: adding
the instance to an existing child, creating a new child, splitting
a child, or merging two children.
\end{lpost}

\vskip 0.02in
\noindent
The first option does not revise taxonomic structure; it simply 
updates probabilities for the child. The second alternative creates
a new child based on the instance. The latter two actions lead to
local restructuring before sorting continues. The same evaluation
criterion used in routing determines which of these steps to take.
Because Cobweb learns incrementally, the order in which it encounters
instances can influence the structure of the acquired taxonomy.
Merging and splitting can mitigate these order effects, resulting 
in hierarchies with better organizations.

The framework also invokes a deterministic restructuring action
when an instance reaches a terminal node N. In these cases, the
learning process creates two new terminal nodes, one based on the
stored instance and the other based on the new case. The existing
node N becomes their parents, with its probabilities updated based 
on values in the recent instance. Together, these operations 
support the incremental and unsupervised acquisition of a taxonomy 
of probabilistic concepts, which are key characteristics of human 
category learning. 


\section{A Unified Theory of Chunking and Categorization}

Cobweb unifies ideas from decision trees, naive Bayesian classifiers,
and nearest neighbor in an elegant framework, and it makes contact
with the psychology literature on categorization and concept learning.
However, it does not explain the representation, organization, use,
or acquisition of chunks, which are equally important facets of
human and machine cognition. The issue is that Cobweb has nothing
to say about {\it composition\/}, which is central to chunks and their
processing. This section introduces new postulates that, when combined
with those just described, provide an extended theory that links
concepts and chunks. We use examples from language syntax to clarify 
the ideas because they require both capabilities, although the
framework has much broader implications.


\subsection{New Representation Postulates}

The extended theory retains Cobweb's representational assumptions,
including the distinction between instances and concepts, but adds
postulates that make it richer. These include:

\vskip 0.05in

\begin{rpost}
An {\it experience\/} is a set of elements and the local relations 
among them.
\end{rpost}

\vskip 0.02in
\noindent
For example, someone reading text encounters a set of words that
are arranged in a linear sequence. The words are elements that
are related by a {\it before\/} or a {\it left-of} relation.

\vskip 0.05in

\begin{rpost}
Every element is either a {\it primitive\/} structure or a {\it                 
composite\/} structure.
\end{rpost}

\vskip 0.02in
\noindent
Primitive elements are the basic building blocks of memory, similar 
to instances and concepts in Cobweb. A composite, in contrast, is an
element built from primitives or lower-level composites, which its
content description specifies. Composites correspond to {\it chunks\/}.

\vskip 0.05in

\begin{rpost}
Each element can be described by its {\it content\/} or its
{\it context\/}.
\end{rpost}

\vskip 0.02in
\noindent
The first of these specifies an element's structure in terms of its
constituents (e.g., {\sl black cat}). The second specifies other
elements in its vicinity (e.g., {\sl the}, {\sl small}, {\sl chased}),
along with their relations to the element (e.g., {\sl before} or
{\sl after}). We will refer to these as {\it content instances\/} 
and {\it context instances\/}. 



\subsection{New Organization Postulates}

The expanded theory also retains Cobweb's commitment to organizing
concepts in taxonomies, but structures composed of others require
further assumptions. Taken together, the content fields of composites
impose an implicit {\it partonomic\/} hierarchy, and the distinction
between content and context instances leads to additional
organizational postulates. These include:

\vskip 0.05in

\begin{opost}
Long-term memory contains a {\it content taxonomy\/} that organizes
concepts by their contents.
\end{opost}

\vskip 0.02in
\noindent
For primitive elements, this hierarchy corresponds to that in Cobweb,
which clusters instances that have similar attribute values. However,
for composite elements, it groups cases that have similar sets
of constituents. For example, the chunks {\sl black dog} and {\sl 
black cat} share the element {\sl black}. 

\vskip 0.05in

\begin{opost}
Long-term memory contains a {\it context taxonomy\/} that organizes
concepts by their contexts.
\end{opost}

\vskip 0.02in
\noindent
This hierarchy instead organizes concepts that occur in similar
contexts. For instance, it might group the words {\sl dog} and
{\sl cat} together because they appear in like surroundings.
The same applies to composite structures, like different
types of noun phrases, which occur both before and after verbs.
These two taxonomies are intertwined in that nodes in one point
to nodes in the other. However, keeping them separate acknowledges 
that one can organize memory either in terms of content (i.e.,
the constituents of chunks) or in terms of context (i.e, the
situations in which chunks occur).



\subsection{New Performance Postulates}

Performance in Cobweb focuses on classification and prediction,
but the introduction of chunks supports new types of cognitive
activities. In language processing, these are referred to {\it                  
parsing\/} and {\it generation\/}, and the extended theory makes
commitments to their operation as well.




\vskip 0.05in

\begin{ppost}
Parsing iteratively creates a partonomic tree from the bottom up,
adding chunks that elaborate an input experience.
\end{ppost}

\vskip 0.02in
\noindent
In language processing, this involves creating a parse tree from
the words in a sentence by introducing nonterminal symbols like
{\sl noun}, {\sl noun phrase}, and {\sl verb phrase}.

\vskip 0.05in

\begin{ppost}
Parsing sorts candidates through both taxonomies and selects the best
recognized chunk.
\end{ppost}

\vskip 0.02in
\noindent
This requires a notion absent from Cobweb, a {\it recognition
threshold\/} that a candidate must exceed at some node in the taxonomy
before counting as familiar enough to proceed.\footnote{This idea is 
similar to Fisher's (\citeyear{fisher1989noise}) notion of a prediction 
frontier, in which the taxonomy marks nodes to use when inferring 
values. However, our theory instead assumes a frontier of nodes 
that are marked as recognizable for purposes of creating chunks, 
not for making predictions.} Among the candidates that pass, the 
best-scoring chunk joins the growing parse tree.

\vskip 0.05in

\begin{ppost}
Parsing halts when no candidate is recognized or when only one
composite element remains.
\end{ppost}

\vskip 0.02in
\noindent
The content score asks whether the composition looks like one used
before; the context score asks whether the surroundings look like
ones that have hosted similar composites. The termination condition
allows graceful failure: when nothing clears the threshold, parsing
ends ends rather than asserting spurious chunks at the boundary of
competence.

The generation process also moves beyond Cobweb's capabilities. This
proceeds in the opposite direction from parsing with the aim of
creating a well-formed collection of new elements.

\vskip 0.05in

\begin{ppost}
Generation iteratively expands a composite element from the top down,
decomposing chunks to produce an experience.
\end{ppost}

\vskip 0.02in
\noindent
This is the inverse of parsing. Rather than combining elements into
larger structures, it starts from a single composite and rewrites it
into constituents, then rewrites those in turn until only primitives
remain. For language, expanding a {\sl sentence} into a {\sl noun
phrase} and {\sl verb phrase}, decomposing them, and so forth, 
yields a grammatical sequence of words.

\vskip 0.05in

\begin{ppost}
Generation recalls candidate decompositions from the content taxonomy
and conditions them on the context taxonomy.
\end{ppost}

\vskip 0.02in
\noindent
This assumes a counterpart to the recognition threshold, which
determines whether a candidate is familiar enough to use in parsing.
Here we need a {\it recall threshold\/} that determines whether 
to use a content concept for an  expansion~\citep{gupta2025hierarchicalsemanticretrievalcobweb}. If this is too low, then generation
will draw on very specific nodes supported by only a few cases; if
too high, it will favor general nodes that mix constituents
which do not belong together. Recall samples from the most specific
chunks that exceed the threshold to generate candidate expansions,
then uses the context hierarchy to rank them.


\vskip 0.05in

\begin{ppost}
Generation terminates when every expanded element is a primitive.
\end{ppost}

\vskip 0.02in
\noindent
The two taxonomies play complementary roles. The content
hierarchy specifies candidate chunk expansions, whereas the context
hierarchy specifies how well they fit their surroundings. For
grammatical expertise, the first is analogous to individual rewrite
rules, such as {\sl VP => Verb NP}, and the second to knowledge about
which rule to choose. Because generation recombines familiar parts in
new ways, it can produce experiences the system has never seen yet
are consistent with what it has learned, which for language means
grammatical sentences beyond the training corpus. 

\newpage

\subsection{New Learning Postulates}

Learning in the extended theory does not require new postulates about
mechanisms, as they remain unchanged from the Cobweb framework. 
However, it does make claims about the inputs to these already 
established processes. They include:

\vskip 0.05in

\bull
Learning incorporates content and context instances extracted from 
elements into the content and context taxonomies, respectively.

\bull
Learning incorporates instances for composite elements into both
taxonomies, but incorporates instances for primitive elements only 
into the context taxonomy.

\vskip 0.05in
\noindent
Neither statement should be surprising, as both follow from two other
commitments: (a) the distinction between content and context
instances, and (b) interleaving learning with performance. There is no
separate mechanism for creating chunks, since parsing already produces
candidates, and learning simply treats them as data for updating the
taxonomies. 

\begin{figure}[t]
\centering
\includegraphics[width=\textwidth]{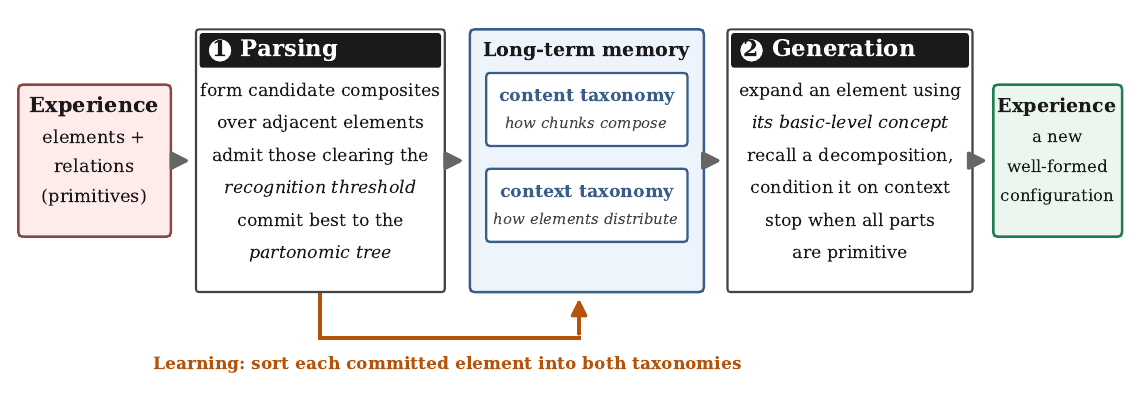}
\vspace{-0.40in}
\caption{Modules of \trellis/ and the flow of an experience through
  them. Parsing composes primitive elements into a partonomic tree,
  consulting the content and context taxonomies that make up long-term
  memory, while generation inverts the process. Section~3 defines 
  the terms set in italics.} 
\label{fig:architecture}
\vskip 0.15in
\end{figure}

This does not reflect a limitation of Cobweb's account of learning.
If anything, it provides evidence for substantial generality. Cobweb
was designed to learn from independent and identically distributed
data, which sequential domains like language clearly violate.
Yet combining its original learning methods with the distinction
between content and context instances, it can cover types of
phenomena for which its creator did not intend.

\section{\trellis/: An Implementation of the Theory}

The extended theory offers an account that unifies concepts and
chunks, but it is abstract. To make it operational, we have developed
\trellis/, which embeds the postulates in an implemented system that
invokes Cobweb as a subroutine. In this section, we describe how
\trellis/ elaborates the theory and illustrate its choices with
examples from language processing. We delay covering details like
formulas for the recognition score until Appendix~B.

Figure~\ref{fig:architecture} summarizes how the pieces fit together 
and clarifies our terminology. An experience enters as a set of 
primitive elements and relations among them. Parsing composes these 
into higher-level elements, admitting only candidates that clear 
a recognition threshold and committing the best to a partonomic tree.
Long-term memory holds two taxonomies, over content and over context
descriptions, which supply the scores that parsing consults. Generation
runs the same machinery in reverse. Learning is not a separate stage:
every committed element is sorted into both taxonomies, so the memory
that scores later experiences is the one built from earlier experiences.

\subsection{Representation in \trellis/}

\begin{figure}[t]
\centering
\includegraphics[width=\textwidth]{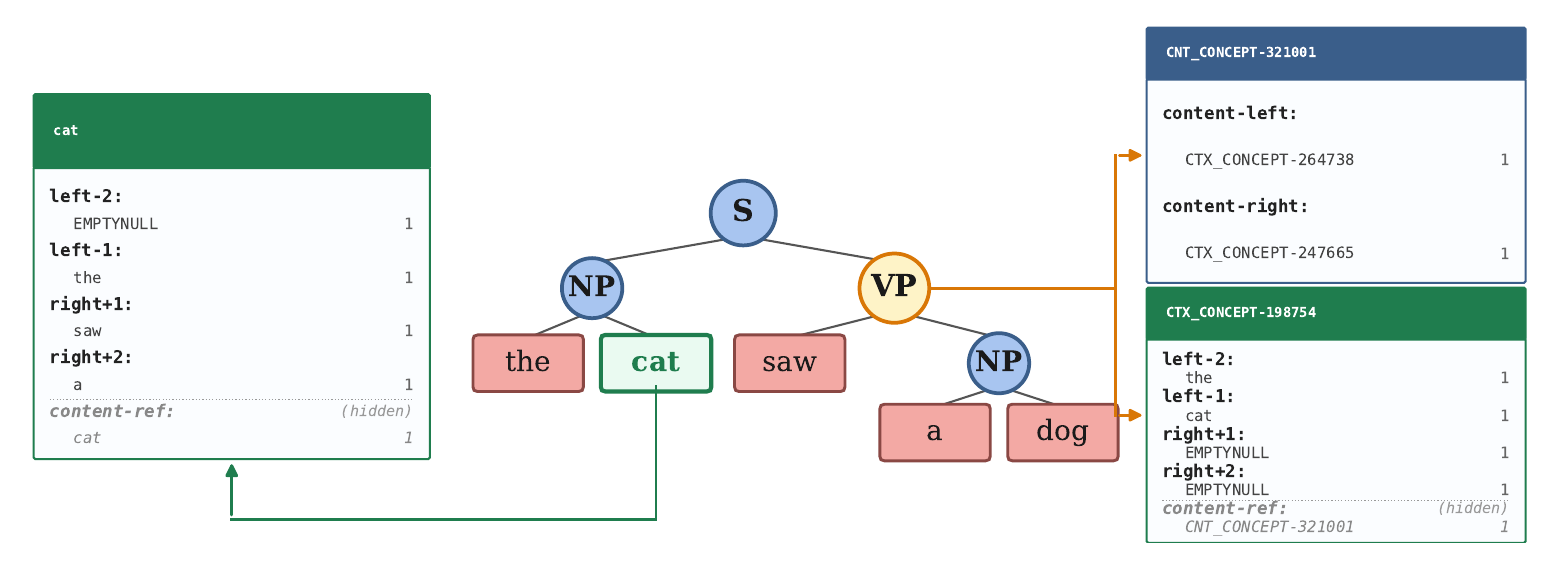}
\vskip -0.10in
\caption{One primitive (``cat'', green) and one composite (``saw a
  dog'', amber) from a parse of ``the cat saw a dog''. Each box header
  is the long-term memory identifier to which the instance was sorted. 
  The composite's content field
  points into the context hierarchy, keeping the two descriptions
  in complementary formats. Window width, distance weighting, and
  content-bag width are run-time values (Appendix~B).}
\label{fig:instances}
\vskip 0.10in
\end{figure}

Every element in \trellis/ includes two fields: {\it content\/}
specifies the parts that compose the element, whereas {\it context\/}
names its neighbors~\citep{matsakis-cobweb}. Primitives have only a context field, while
composites have both. Figure~\ref{fig:instances} illustrates one
primitive and one composite drawn from the parse of a small sentence.
A composite's content field encodes its left and right children
with two attributes: an identifier that points to the context
taxonomy (referring to the child itself) and a complexity tag
(a small integer denoting its nesting depth). Both reside in the
context hierarchy's identifier space rather than the surface
vocabulary, so the content hierarchy never sees a primitive word.
Every value encountered during a content sort was created by a
context sort of the corresponding child. The complexity tag lets
the content taxonomy identify composites whose children have
similar contexts but differ in structural depth, such as
{\sc np}$=${\sc det}$+${\sc n} vs.\ {\sc s}$=${\sc np}$+${\sc vp}.

\subsection{Organization in \trellis/}

\begin{figure}[t]
\centering
\includegraphics[width=0.95\textwidth, trim=0 0 15 15, clip]{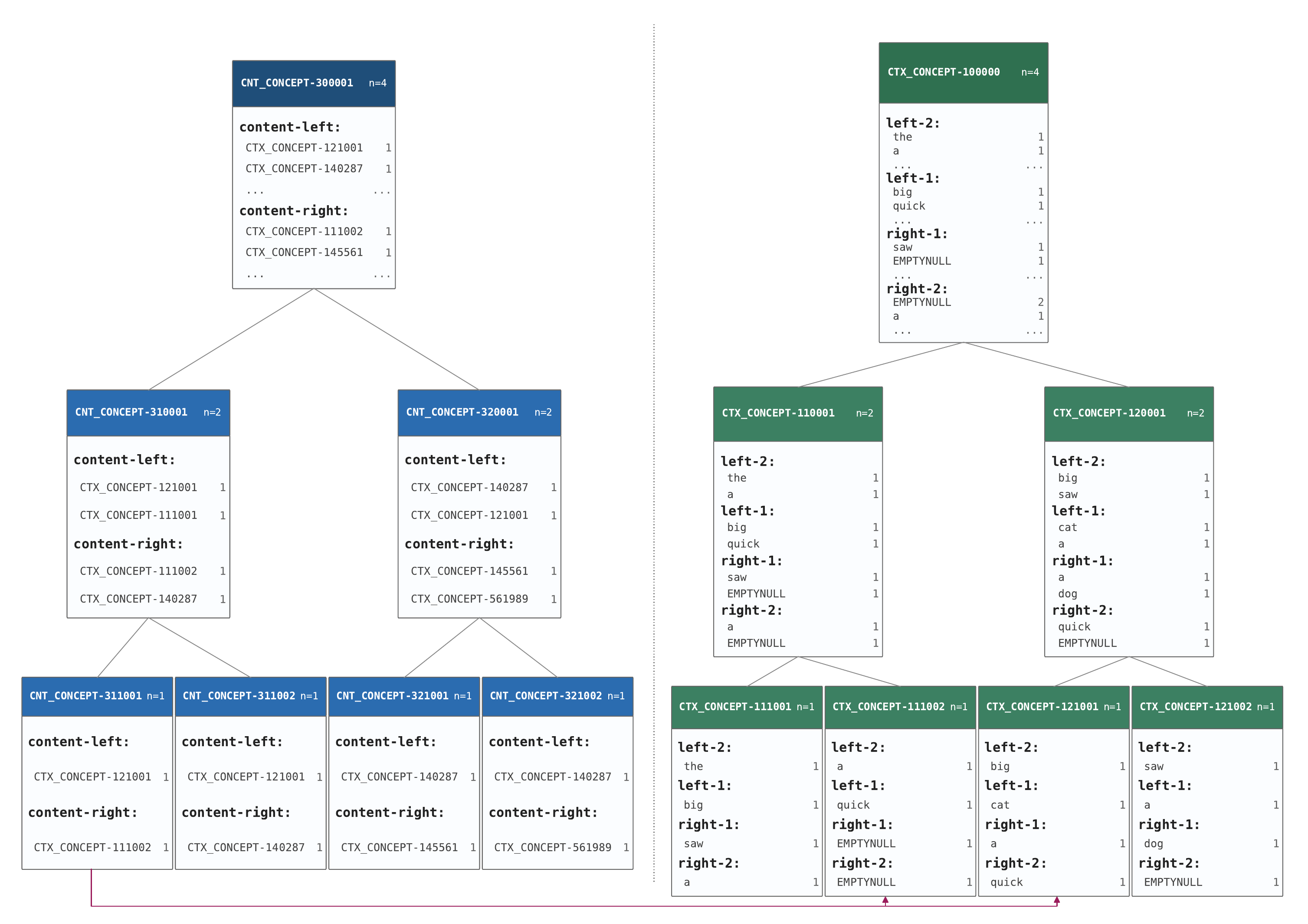}
\vskip -0.07in
\caption{Leaf-level subtrees of the content (left) and context (right)
  hierarchies. Arrows show the cross-hierarchy link: a content leaf's
  child fields name the context concepts that describe its children.}
\label{fig:hierarchies}
\vskip 0.15in
\end{figure}

Following the extended theory, \trellis/ divides long-term memory into
two Cobweb taxonomies. The content hierarchy stores the content
descriptions of composites and the context hierarchy stores the
contexts in which they occur. They are linked in that a composite's
content refers to identifiers in the context hierarchy, so every
description it encounters comes from an earlier context classification. 
Neither trades conceptual facets for compositional ones, since both
structures carry each. Composites with structurally similar children
appear near each other in the content tree, reflecting a shared role
(say, as an NP-filling constituent),
while a noun and an adjective phrase appear near each other in the
context tree when they occur in comparable neighborhoods.
Figure~\ref{fig:hierarchies} shows leaf-level subtrees of each
hierarchy and the cross-hierarchy references that link them.



\subsection{Parsing and Generation in \trellis/}

\begin{figure}[t]
\centering
{\small (a) One parsing step}\\[0.10in]
\includegraphics[width=0.95\textwidth]{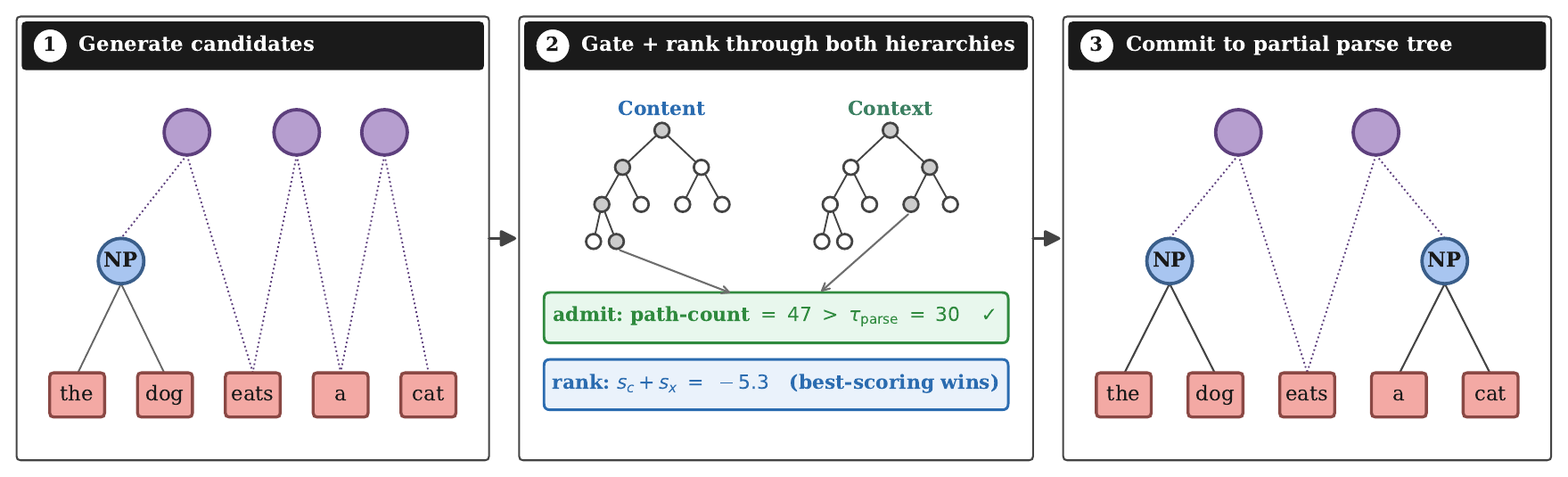}\\[0.04in]
{\small (b) One generation step}\\[0.10in]
\includegraphics[width=0.95\textwidth]{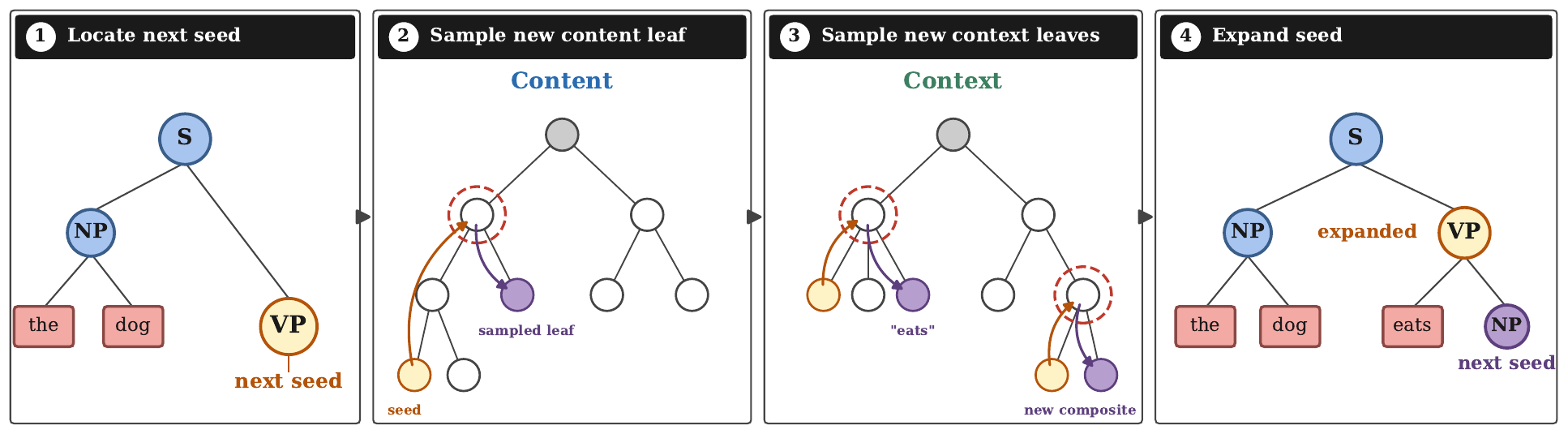}\\[0.04in]
\vskip -0.05in
\caption{{\sc Trellis}' two performance modules. (a) Parsing (1) 
  forms candidate
  chunks over adjacent frontier elements; (2) sorts each through both
  taxonomies, admitting it iff some nonroot ancestor on the sort path
  has count above $\tau_{parse}$, ranking the admitted candidates by
  $s_c + s_x$; (3) commits the top-ranked one to
  the parse tree. (b) Generation (1) selects the next seed; (2) samples
  a fresh content leaf from a well-supported neighborhood above it; (3)
  resolves each of the sampled leaf's two child references against the
  class the parent expects; (4) expands the seed with these results.
  The numeric values in a2 are for illustration only.}
\label{fig:parse-infographic}
\label{fig:generation-infographic}
\vskip 0.2in
\end{figure}

The \trellis/ system includes two modules that expand on the
theory's performance postulates. 
As Figure~\ref{fig:generation-infographic}~(a) depicts, a parsing 
mechanism constructs parses from the bottom up by iteratively 
creating chunks that elaborate on the input experience. 
In addition, the generation mechanism shown in 
Figure~\ref{fig:generation-infographic}~(b) constructs output 
experiences from the top down by iteratively decomposing chunks 
until its reaches primitive elements.

Parsing carries out greedy search through the space of possible
parses. The module starts with a frontier of primitive elements,
such as the words in a sentence, and repeatedly proposes candidate
chunks over adjacent pairs. It sorts each candidate's content and
context descriptions through their respective hierarchies, scoring
them by how well the candidate matches the concept at which sorting
halts relative to its competitors (Appendix~B). \trellis/ sums the
two scores to produce a combined recognition score. For
``the cat saw a dog'', the first step would consider four
adjacent word pairs.

\trellis/ admits a candidate only when a concept through which it
was sorted has been visited enough times, with primitives gated on
the context hierarchy and composites on the content hierarchy. The
module walks upward from the node at which sorting halted; the
candidate is recognized if the count of some nonroot ancestor
exceeds the recognition threshold $\tau$. This gating scheme
suppresses selection early in learning but has little effect later.
After this, \trellis/ scores candidates by their posterior probabilities 
based on nodes in the two hierarchies and replaces the selected 
candidate's constituents with the composite element on the frontier. 
Parsing terminates if the module recognizes none of the candidate 
chunks, in which case it halts with a partial parse.

Generation proceeds in the opposite direction to produce a
grammatical structure. This process begins from a seed node that
is an unexpanded composite with an identifier in the content
hierarchy. The module samples from a {\it recombination pool\/}
associated with the seed, decomposing the result into two children.
To create the pool, \trellis/ walks up from the seed's leaf until
it reaches an ancestor whose count exceeds the recall threshold 
$\tau$, analogous to the recognition threshold. At this point, 
it considers the constituent pairs stored at this node, each of 
which has an associated probability. Thus, the system draws from a 
set of decompositions it has seen often enough to treat as acceptable, 
from which it then samples, rather than ones with little empirical
support. The sampling process can produce new combinations that
no single training experience has contained.

After selecting a decomposition, \trellis/ uses the context hierarchy
to filter alternatives. If a candidate's parent does not expect one 
of them, then the system eliminates it from competition to ensure it
cannot emit an incorrect constituent. If the filter eliminates all
candidates, say because the context concept has not yet acquired any
compatible chunks, then the system falls back to a wider, less-filtered
pool that incurs a small risk for errors of commission in return
for broader coverage. Generation halts when every leaf of the parse
tree is a primitive structure.

\subsection{Learning in \trellis/}

Learning in \trellis/ does not require any new mechanisms beyond those
in Cobweb, but it operates over what we have called `experiences',
which are collections of related elements. We provide system 
not only with training experiences, but also with desired parse trees
for them. For example, the sentence {\sl the big dog chased the black 
cat} would come with a parse into phrases at various levels. This
constitutes a form of supervised learning, although nonterminal nodes
in the parse trees are not themselves labeled with terms like {\sl NP} 
and {\sl VP}. 

As with parsing, \trellis/ learns iteratively from the bottom up.
This involves selecting an element on the frontier (initially the
set of primitive structures) and sorting it through taxonomies.
The system only sends primitive elements through the context
hierarchy, as they have no internal structure. Composite elements
are sent through both taxonomies, which incorporates them by their 
content and context descriptions. \trellis/ classifies composite
elements only after it has already processed their constituents.



Once the system has handled every element of a parsed experience, it
turns to the next, continuing until it exhausts the training data.
Thus, learning occurs both incrementally within each experience and
across a succession of experiences. \trellis/ has no separate routine
for creating chunks: every composite structure becomes a new node in
the taxonomies, and that node is what later parsing uses as the 
category label when the system meets a structurally similar chunk. 

\section{Evaluation of \trellis/}

We maintain that \trellis/ is a faithful implementation of our unified
theory, but whether it operates as intended is another question. To
answer it, we designed experiments on grammar acquisition that examined
its ability to represent, organize, use, and learn linked content and
context taxonomies. We describe the experimental design and test
domains below, then present the results. As noted earlier, we focused 
on syntax because it requires both chunks and categories, but we 
anticipate our findings will carry over to other settings. 


\subsection{Experimental Design}

The acquisition of grammatical expertise is harder to study than
learning for classification tasks. We follow Langley and Stromsten's
(\citeyear{grids-langley-stromsten}) methodology, which recorded two
dependent variables: {\it errors of omission\/} and {\it errors of
commission}. The first measures how far a learned grammar is overly
specific, in that it fails to parse or generate some legal sentences
with correct parses. The second measures overgeneralization, in that
the grammar parses or generates illegal constructions.

For our initial studies, we devised synthetic context-free grammars
to generate training data, rather than a corpus of natural sentences
like the Penn Treebank. To compute the two metrics, we used the target
grammar to generate training sentences and associated parse trees,
then let \trellis/ learn concepts and chunks from these experiences.
To measure errors of omission, we used the target grammar to generate
novel sentences, then used \trellis/' parsing module to attempt
parsing them. Similarly, to measure errors of commission, we called
the generation module to produce sentences from the learned taxonomies,
then used the target grammar to attempt parsing them.
Because the system can acquire incomplete grammars, especially early
in training, we did not use an all-or-none measure of success.
Instead, we gave partial credit for each substructure of the target
parse. In particular, we compared the {\it span\/} of words covered
by each subtree,\footnote{The grammar induction literature 
typically calls such spans {\it brackets\/}; we use {\it substructure\/}
to avoid the Penn Treebank's labelled-bracket convention, which
combines structural span agreement with part-of-speech or phrasal
labels.} as \trellis/' memory has no class labels on nonterminals.
For each test sentence, we took the fraction of substructures matched
between the target and learned sentences.

The main purpose of our experiments was to reveal whether \trellis/
induces accurate grammatical knowledge or whether it forms overly
specific or overly general structures. We were also interested in
the rate of improvement as measured by learning curves. In addition,
we also wanted to know how behavior scales in response to two types
of grammatical complexity. One involved the number of nonterminal
categories, such as {\sl NP} and {\sl VP}, in the target grammar.
Raising this count increases the number of chunk types that the content
hierarchy must encode and acquire. The other concerned the words per
part-of-speech class. Increasing this number requires the context
hierarchy to generalize across more surface forms.



In one experiment, we varied the number of nonterminal symbols, using
three grammars with three, six, and eight nonterminals, which we will 
refer to as small, medium, and large, while holding number of words 
per class constant. In a second study, we varied the number of words 
per part of speech class from 11 (low) to 22 (medium) to high (39), 
while holding the number of nonterminals at six. Appendix~A presents 
the rewrite rules and lexical items for each grammar, most of which
involve recursion, while Appendix~B provides the system parameters,
which we did not vary. Each experimental condition used 400
sentence-parse pairs generated from the target grammar. To measure
errors of omission, we used five-fold cross validation, holding out
a different subset of 80 pairs for each fold in testing and the
remaining 320 pairs in training. To measure errors of commission, we
used the same five learned taxonomies to generate 80 sentence-parse
pairs. Finally, we averaged each dependent variable over five runs based
on the different training sets.





\subsection{Experimental Results}

\begin{figure}[t]
\centering
{\small (a) Varying grammar complexity}\\[0.10in]
\includegraphics[width=0.95\textwidth]{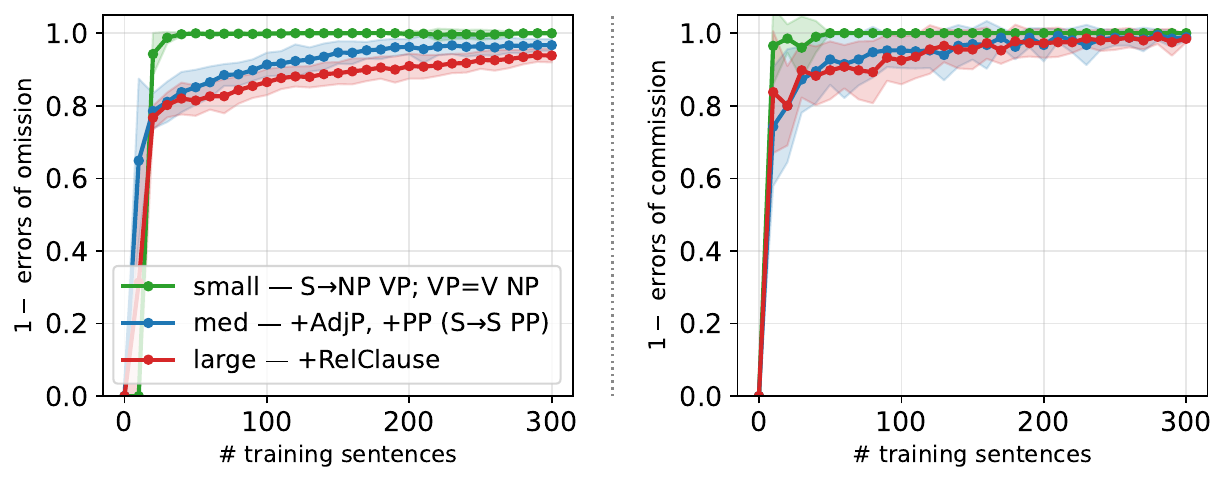}\\[0.04in]
{\small (b) Varying lexicon size for the {\sc med} grammar}\\[0.10in]
\includegraphics[width=0.95\textwidth]{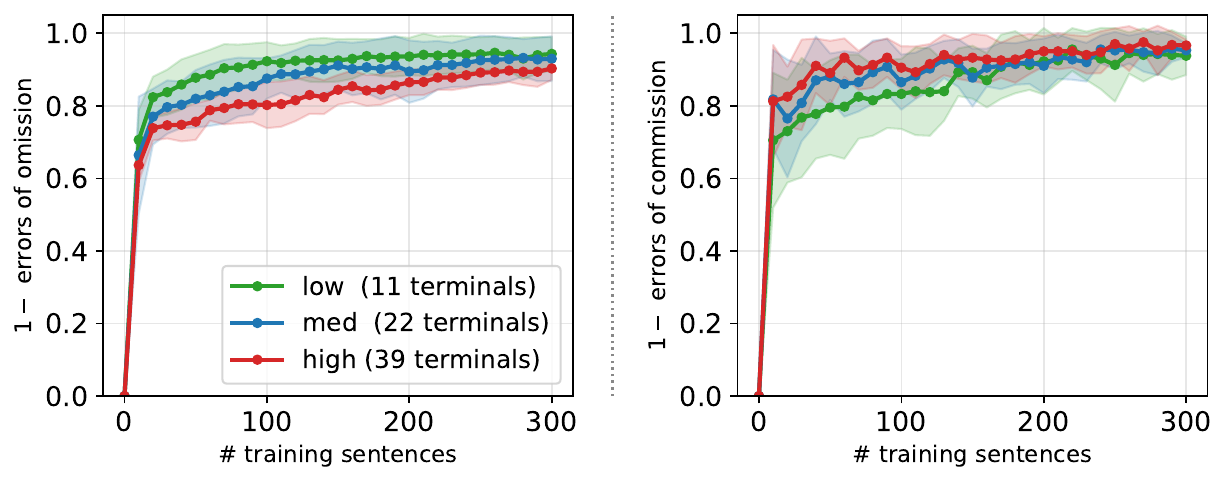}\\[0.04in]
\vskip -0.10in
\caption{$1 - {\rm omission\ rate}$ (parse, left of each pair) and
  $1 - {\rm commission\ rate}$ (generation, right of each pair), across
  (a) the three grammars and (b) the three lexicon sizes (five data
  sets, $\pm1\sigma$ band).}
\label{fig:comparison-grammar}
\label{fig:comparison-terminal}
\vskip 0.1in
\end{figure}

Figure~\ref{fig:comparison-grammar}(a) shows the results of the first
study, plotting one minus the omission rate on the left and one minus
the commission rate on the right against the number of training
sentences. Each graph shows three learning curves, one per grammatical
complexity level. The most important result is that \trellis/ acquires
grammatical expertise well, with both errors of omission and commission
dropping to low percentages even for the most complex grammar, despite 
the inclusion of recursive rewrite rules in the targets. 
Learning is also rapid, with both errors dropping to 20 percent almost
immediately and then improving gradually with more training sentences.
Increasing the number of nonterminals slows learning, as the system
requires examples of every chunk type to master the grammar, but the
degradation is graceful. Inspection of the content hierarchy showed
reasonable concepts for each nonterminal in the target grammar.


The plots in Figure~\ref{fig:comparison-terminal}(b) reveal similar
results from the experiment that varied the lexicon size. As
before, \trellis/ comes close to mastering the grammars from
a moderate number of training sentences, with low errors of
omission and commission for all complexity levels. Increasing
the number of words per part of speech class slows the rate of
learning, but improvement is rapid in all conditions and the
reduction with more complex grammars is modest. Inspection of
the context hierarchy, which encodes word and phrasal classes, 
showed the system grouped nouns with nouns, verbs with verbs, 
and so forth, as desired.


In general, both experiments produced results consistent with our
expectations. They showed that \trellis/ can acquire accurate
grammatical expertise, with low errors of omission and commission,
from a moderate number of sentence-parse pairs. The rate of learning
was rapid, but increasing the number of nonterminal symbols and
lexical items led to modest slowing, suggesting that the system
scales reasonably in response to both forms of grammatical complexity.
Examination of the content hierarchy showed concepts that mapped well
onto nonterminals in the target grammar and inspection of the context
hierarchy showed categories for target word classes like determiners,
adjectives, nouns, verbs, and prepositions. 

Overall, these results indicate that \trellis/' extension of Cobweb
to support linked content and context hierarchies support the rapid
acquisition of chunks and concepts, both of which are important
in grammar induction. We considered briefly the idea of a lesion study
that compared \trellis/' behavior to that of Cobweb, but rejected it
because the earlier system only handles individual items and not
collections of related items like words in sentences. The experiments
demonstrate clearly that the new system's ability to represent, use,
and learn both concepts and chunks give it this ability. Certainly
many other systems can acquire grammatical knowledge, but few of
them offer such a unified theoretical account of chunks and concepts.

\begin{figure}[t]
\centering
\includegraphics[width=0.95\textwidth]{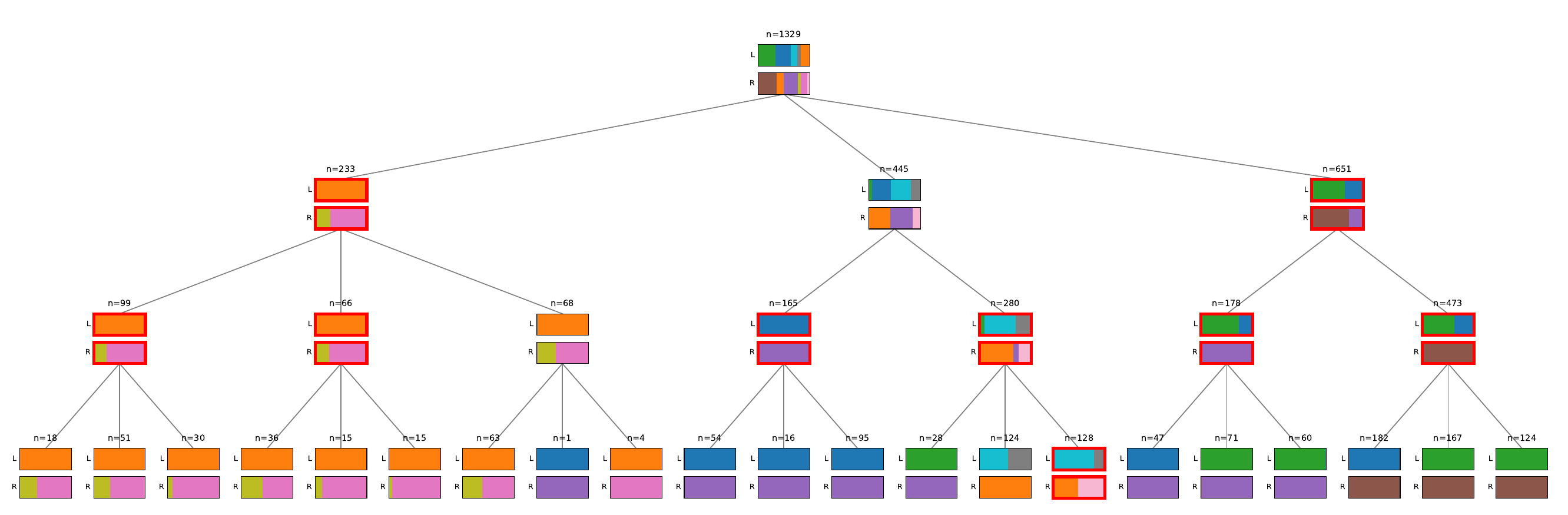}\\[0.03in]
\includegraphics[width=0.95\textwidth]{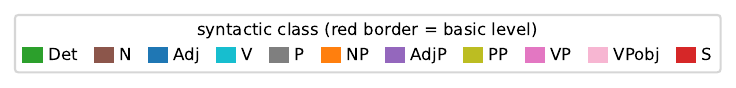}\\[0.03in]
\includegraphics[width=0.95\textwidth]{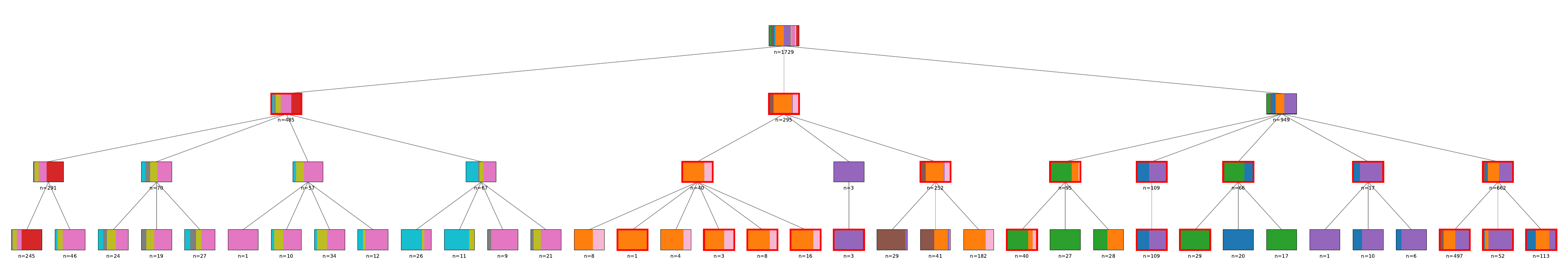}
\caption{Content (top) and context (bottom) taxonomies after training
  on the \textsc{med} grammar. Each node shows the distribution of
  syntactic classes among the elements sorted through it, with $n$ the
  number of such elements. For content nodes the upper and lower bars
  give the left and right constituent, respectively. Red borders mark basic-level nodes, and the legend applies to both panels. The labels 
  are only for inspection purposes and play no role in learning. Nodes 
  become nearly pure well above the leaves, which is the generalization 
  that lets one chunk cover many fillers.}
\label{fig:hierbars}
\vskip 0.20in
\end{figure}

The aggregate scores say that \trellis/ parses and generates well, but
not what it has learned. Figure~\ref{fig:hierbars} opens the two
taxonomies after training on the \textsc{med} grammar, coloring each 
node by the syntactic classes of the elements beneath it; the labels 
are ours for interpreting the structure, not used by the system. 
Two patterns stand out. First, 
nodes are close to pure well above the leaves, so a node stands for a
class of elements rather than the items that created it. Second, the two
taxonomies carve memory along different dimensions: content separates
chunks by the classes of their parts, splitting determiner-noun pairs
from verb-phrase structures, whereas context groups elements appearing
in similar surroundings, placing nouns beside adjective phrases that
fill the same slots. This division of labor is what lets a chunk learned
over one set of words apply to another. 

\section{Related Research on Concepts and Chunks}

Our unified theory offers a promising account of how concepts and
chunks are two sides of the same coin and the \trellis/ results
suggest its viability, yet other frameworks have addressed the problem
and merit discussion. The closest are other variants of Cobweb,
starting with \textsc{Labyrinth} \citep{labyrinth-thompson-langley}, which
encoded, used, and learned both primitive and composite concepts, the
latter similar to chunks. However, it lacked a contextual taxonomy and
suffered substantial order effects. Another descendant, Trestle 
\citep{trestle-maclellan}, operated over instances organized
in a partonomic hierarchy, but flattened the representation during
processing. A convolutional version of Cobweb \citep{convolutional-cobweb} learns over multiple levels of image descriptions, giving 
some effects of chunks without treating them explicitly.


A second line of Cobweb research treats problem solving as the source 
of composite structures. 
Yoo and Fisher's (\citeyear{yoo-fisher-explanations}) EXOR system 
formed concepts over successful problem decompositions, which
had forms much like our parse trees, but it generated these from 
domain rules provided in advance. In addition, EXOR constructed a 
single taxonomic hierarchy, with each node storing both a chunk-like 
decomposition and a probabilistic summary used to retrieve it. Other 
work by Yang and Fisher (\citeyear{yang-fisher-plans}) and by Langley 
and Allen (\citeyear{langley1993unified}) reported systems that used 
Cobweb-like methods to cluster operators from successful means-ends 
problem solving, but their taxonomies did not store explicit 
decompositions. In addition, Fisher and Pazzani 
(\citeyear{fisher-pazzani-theory-guided}) argued 
that clustering should be guided by the function that categories 
will serve, which is one way to read our decision to maintain 
separate content and context taxonomies.

Research on discrimination networks shares many features with the
Cobweb family. EPAM \citep{feigenbaum-epam, epam-richman} organizes 
knowledge in a univariate
taxonomy that it learns with incremental, unsupervised methods, and
its operations for extending the network are the same as Cobweb's
but predated them by 25 years. The CHREST architecture 
\citep{chrest-gobet-lane}
extends EPAM on multiple fronts, including the addition of lateral
links among nodes, similar to the interleaving in \trellis/. Moreover,
the framework has always supported using terminal nodes (e.g., letters) 
as tests for recognizing higher-level discriminants (e.g., words). 
But despite similar treatments of categorization, their accounts of
chunking follow different paths. In particular, EPAM and CHREST 
parse structures from the top down, whereas \trellis/ operates 
from the bottom up. 

Although seldom recognized, neural networks \citep{rumelhart-backprop} also offer a
unified narrative about concepts and chunks. Nodes in such a network
have very different semantics from those in Cobweb and EPAM, but we
can still view them as graded concepts that match to a greater or
lesser degree. But we can also view them as chunks that
combine simpler elements, as demonstrated by visualizations of learned
`features' as weight maps. Taken together, this framework would offer
a compelling unification of concepts and chunks if only it included
operations for creating new structures, rather than merely updating
weights, did not suffer from catastrophic forgetting, and supported 
more rapid, human-like learning.

A fourth paradigm, with fewer links to cognitive science, involves
the construction of context-free grammars from training cases.
Classic systems in this paradigm \citep{grids-langley-stromsten, 
wolff-grammar, wolff2026intelligence} include operations for
inventing nonterminal symbols (creating chunks) and merging them
(forming classes). Although most such systems are strongly committed
to symbolic representations, the learned structures are organized in 
what Langley (1996) has called {\it inference networks\/} that are 
similar in spirit to neural networks. Research on predicate invention
for inductive logic programming \citep{muggleton-buntine-predinvention}
has championed related ideas for introducing chunk-like terms. 
Despite many differences from Cobweb and EPAM, these lines of work 
have been important sources for our ideas about concepts and chunks. 

Our theory also bears a resemblance to some frameworks for logical
representations of knowledge. For instance, the KL-ONE system
(Brachman \& Schmolze, \citeyear{brachman1985overview}) and its 
descendants organized concept definitions into {\it structured 
inheritance networks\/}.\footnote{We should also note that McCarthy's 
(\citeyear{mccarthy1993notes}) notion of context specifies 
logical conditions on meaning, as opposed to the probabilistic
contexts in \trellis/, which are more closely related to
distributional semantics.} These have much in common with
discrimination networks, with each node being a specialization of its
parents and a classification process sorting descriptions through the
taxonomy. An important difference is that KL-ONE nodes can have
multiple parents whose children denote their composition, which
\trellis/ addresses with separate hierarchies. Another is
that there has been little work on learning structured inheritance
networks from experience. However, in many ways the two frameworks 
are kindred spirits. 

\section{Concluding Remarks}

In this paper, we presented a unified computational theory of concepts
and chunks. We reviewed the assumptions behind Cobweb, which forms a
taxonomy of probabilistic concepts that it uses for classification and
prediction. Next we introduced additional postulates that extend the
framework to support both categorization of experience and composition
into chunks. The extended framework assumes two linked taxonomies, one
based on content descriptions and another on contextual ones. We also
covered {\sl Trellis}, an implementation of the theory that invokes
Cobweb as a subroutine and tested the system on its ability to learn
grammatical expertise from sentence-parse pairs. 
Our theory bridges the longstanding chasm between accounts of concepts
and chunks, which are usually treated as disjoint phenomena. Moreover,
experimental results with \trellis/ suggest not only that it can acquire
syntactic knowledge, which involves both categories and chunks, but
does so from a modest number of training sentences and handle recursive 
structures. 


The creation of chunks is a compression device. Once \trellis/ has 
introduced a concept for some configuration, the system can refer to 
it with a single symbol, so the context it must carry shrinks from 
a large list of primitives to a handful of composites. This is how  
the two taxonomies support each other: the context description of a 
composite spans the same stretch of experience as the descriptions 
of its parts, but it occupies one slot rather than many. Contemporary 
language models take the opposite approach, retaining every token and 
paying with the quadratic cost of relating them. We do not claim that 
\trellis/ competes with such systems at their current scale. We do
claim that a memory which builds and refers to its own units offers 
a better way to handle context than retaining the entire past, and 
the rapid learning in our experiments shows its benefits.

Despite our progress to date, both the theory and its implementation
have limits that we should address. Here are some open questions 
that future work should attempt to answer: 

\vskip 0.05in

\bull
Can we replace \trellis/' greedy parser with a more powerful one that
reconsiders its choices when they lead it astray, as on garden path 
sentences?

\bull
Can we extend chunk learning to identify constituents from sentences 
alone, without providing sample parse trees, like some earlier systems 
for grammar induction?

\bull
Can \trellis/ represent, use, and acquire not only context-free grammars
but also context-sensitive ones, as suggested by its context taxonomy?

\bull
Can the framework learn in other settings that require both generalization
and composition, such as understanding scenes, generating plans, and
playing games like chess?

\bull
Can \trellis/ acquire syntactic expertise not only from sentences 
generated by synthetic grammars but also from natural corpora like 
the Penn Treebank?

\bull
Can the system enable a new variety of large language model that 
has more interpretable structures, performs more efficiently, and 
learns more rapidly than existing ones?

\vskip 0.05in
\noindent
We hope to address each of these challenges in future research by
continuing to incorporate ideas from earlier efforts and combining
them in novel ways, following the tradition established by other
researchers in the cognitive systems movement.

\begin{acknowledgements}
\noindent
The research reported here was supported by Grant No.\ FA9550-23-1-0580
from the US Air Force Office of Scientific Research, which is not
responsible for its contents. We thank Chris MacLellan, Zekun Wang, 
Xin Lian, Kyle Moore, and friends at the Teachable AI Lab for 
useful discussions.
\end{acknowledgements}

\newpage

{\parindent -10pt\leftskip 10pt\noindent
\bibliographystyle{cogsysapa}
\bibliography{main}

}

\newpage

\appendix

\section{Synthetic Grammars}
\label{app:grammars}

The first experiment in Section~5 used three synthetic context-free 
grammars that we describe here. They are strictly nested, so we 
present them as one set of entries, each tagged with the smallest 
grammar in which it appears (\textsc{s}~$=$~\textsc{small}, 
\textsc{m}~$=$~\textsc{med}, \textsc{l}~$=$~\textsc{large}). 
Every non-terminal expansion is strictly binary and adjective-free 
NPs take use the rule NP $\rightarrow$ Det N, so the grammar never 
produces a unary syntactic node. The \textsc{med} grammar adds adjective 
recursion through an AdjP rule that carries at least one adjective, 
along with prepositional phrases and direct objects. The 
\textsc{large} grammar adds relative clauses through  
Nbar and RelPro nonterminals. Sentence generation samples
from alternative expansions for nonterminals, most of which 
have equal probabilities. Three exceptions keep RelClause sentences 
near 20\% of the \textsc{large} corpus and keep adjective chains short: 
\textsc{med} NP splits 3:2 (adjective-free nouns~/~AdjP); \textsc{large} 
NP splits 6:2:1 (adjective-free snoun~/~AdjP~/~Nbar), AdjP and Nbar each 
split 2:1; and the AdjP recursion decays as $(1/3)^{d}$ with depth~$d$.

\begin{center}\small
\begin{tabular}{@{}l@{\ }l@{\qquad}l@{\ }l@{}}
\toprule
\multicolumn{2}{@{}l}{\it Productions} & \multicolumn{2}{l@{}}{\it Lexicon} \\
\midrule
\textsc{s} & S $\rightarrow$ NP VP                      & \textsc{s} & Det $\rightarrow$ the \(\mid\) a \\
\textsc{s} & NP $\rightarrow$ Det N                     & \textsc{l} & Det $\rightarrow$ this \(\mid\) that \\
\textsc{m} & NP $\rightarrow$ Det AdjP                  & \textsc{s} & N $\rightarrow$ cat \(\mid\) dog \(\mid\) man \(\mid\) woman \(\mid\) park \(\mid\) telescope \\
\textsc{l} & NP $\rightarrow$ Det Nbar                  & \textsc{l} & N $\rightarrow$ boy \(\mid\) girl \(\mid\) teacher \\
\textsc{m} & AdjP $\rightarrow$ Adj N \(\mid\) Adj AdjP & \textsc{m} & Adj $\rightarrow$ big \(\mid\) small \(\mid\) red \(\mid\) quick \(\mid\) lazy \\
\textsc{l} & Nbar $\rightarrow$ N RelClause \(\mid\) AdjP RelClause & \textsc{l} & Adj $\rightarrow$ tall \(\mid\) curious \\
\textsc{s} & VP $\rightarrow$ V NP                      & \textsc{s} & V $\rightarrow$ saw \(\mid\) liked \(\mid\) chased \(\mid\) found \(\mid\) admired \\
\textsc{m} & VP $\rightarrow$ V VPobj                   & \textsc{l} & V $\rightarrow$ carried \(\mid\) read \\
\textsc{m} & VPobj $\rightarrow$ NP PP                  & \textsc{m} & P $\rightarrow$ with \(\mid\) in \(\mid\) on \(\mid\) under \\
\textsc{l} & RelClause $\rightarrow$ RelPro VP          & \textsc{l} & P $\rightarrow$ near \\
\textsc{m} & PP $\rightarrow$ P NP                      & \textsc{l} & RelPro $\rightarrow$ who \(\mid\) that \(\mid\) which \\
\bottomrule
\end{tabular}
\vskip 0.30in
\end{center}


The second experiment in Section~5 holds the \textsc{med} rewrite 
rules fixed and varies only the lexicon, drawing the first $k$
words from a shared pool per part-of-speech class. The variants
are strictly nested, with (\textsc{low} having 11 terminals: 
Det $=$ \{the, a\}, N $=$ \{cat, dog, man\}, Adj $=$ \{big, small\}, 
V $=$ \{saw, liked\}, and P $=$ \{with, in\}. \textsc{med} has 22 
words, adding: N $+$ \{woman, park, telescope\}, Adj $+$ \{red, 
quick, lazy\}, V $+$ \{chased, found, admired\}, and P $+$ \{on, 
under\}. \textsc{high} reaches 39 words by adding: Det $+$ \{this, 
that\}, N $+$ \{robot, apple, book, girl, boy, bird\}, Adj $+$ 
\{ancient, blue, tall, wise\}, V $+$ \{carried, read, knew\}, and 
P $+$ \{near, behind\}.





\newpage

\section{Implementation Formulas}
\label{app:formulas}

\trellis/ includes a number of formulae and system-level parameters 
whose settings influence its performance and learning. We describe 
each of them briefly below. 

\paragraph{Recognition score.}
From postulate R3, the log probability of an instance $\mathbf{i}$ under 
a concept $C$ is 
\[
\log P(\mathbf{i} \mid C) = \sum_{(a, v) \in \mathbf{i}} \log P(v \mid C, a), \qquad P(v \mid C, a) = \frac{n_{C,a,v} + \alpha_a}{n_{C,a} + V_a \alpha_a},
\]
where $n_{C,a,v}$ counts value $v$ for attribute $a$ at $C$, $V_a$ is the value-vocabulary size, and $\alpha_a$ the smoothing constant. The system 
ranks candidates by the \emph{class posterior} $\log P(C \mid \mathbf{i}) = 
\log P(\mathbf{i} \mid C) + \log P(C) - \log P(\mathbf{i})$: normalizing by the instance marginal makes the score discriminative, so non-constituents 
have low scores. The ranking key sums the two posteriors, $s_c + s_x$.

\paragraph{Recognition threshold.}
Parsing must decide whether a candidate chunk is familiar enough to
commit, which \trellis/ determines by asking which level of the 
taxonomy already knows it. When sorting halts at a node, the 
system walks upward and selects the first non-root ancestor whose 
count exceeds $\tau_{\text{parse}}$. 
Primitives are gated on their context-hierarchy path and composites on
their content-hierarchy path, since those are the paths that identify
them. In every reported condition we use $\tau_{\text{parse}} = 30$.
This strategy keeps \trellis/ from using chunks until they have 
been encountered at least a few times, increasing their reliability. 

\paragraph{Recall threshold.}
Generation must also decide how general a chunk to expand, and
\trellis/ answers this with the same upward walk, used twice. First,
the system starts at the seed's leaf and climbs to the deepest ancestor
with a count above $\tau_{\text{sub}}$, which we set to 20. It then
samples a new leaf below that node, giving a different constituent of
the same kind. Second, it climbs to the deepest ancestor with a count
above $\tau_{\text{gen}}$, which we set to 50, and gathers the chunks
stored below it. These form the recombination pool of Section~4.3,
which lets the system combine parts that never appeared together in
training. The lower threshold keeps a substitution within its type,
whereas the higher one widens the pool. Finally, the context hierarchy
rejects any chunk that does not fit the class its parent expects.

\paragraph{Parameter values.}
All conditions in our experiments used the same parameter settings.
The two smoothing constants $\alpha_\text{cnt}$ and $\alpha_\text{ctx}$ in the recognition score decide how
much probability an attribute holds back for unseen values. We set
these to $10^{-4}$ in the content hierarchy and $10^{-5}$ in the
context hierarchy, as the smaller value makes context sorting sharper
and keeps the two taxonomies apart. Each context instance covers five
positions on either side of an element. The nearest neighbor has a
weight of one half, and each step further out has half as much again.
We set the recognition threshold $\tau_{\text{parse}}$ to 30, the
substitution threshold $\tau_{\text{sub}}$ to 20, and the pool
threshold $\tau_{\text{gen}}$ to 50.

\paragraph{Content-instance encoding.}
For each child, the content field of Section~4.1 stores not one
identifier but a bag of up to three concept identifiers from depth four
of the context hierarchy, which preserves the concept's uncertainty
rather than committing to its likeliest value. Because the context
hierarchy restructures as instances arrive, these references can dangle,
so \trellis/ canonicalizes each to a recent ancestor of its target leaf
and rewrites stored instances accordingly.

\end{document}